\documentclass[journal]{IEEEtran}
\usepackage{amsmath}
\usepackage{subfigure}
\usepackage{multirow}
\usepackage{color}
\usepackage{adjustbox}
\usepackage{url}
\usepackage{bbding}
\usepackage{balance}
\usepackage{array}
\usepackage{arydshln}
\usepackage{booktabs}
\usepackage{txfonts}
\usepackage{graphicx}
\usepackage{float}

\usepackage[colorlinks, 
            linkcolor=red, 
            anchorcolor=blue, 
            citecolor=green
            ]{hyperref}

\usepackage[switch]{lineno}
\begin{document}

\title{Efficient Semantic Segmentation via Lightweight Multiple-Information Interaction Network}


\author{Yangyang Qiu,
        Guoan Xu,
        Guangwei Gao,~\IEEEmembership{Senior Member,~IEEE,}
        Zhenhua Guo,
        Yi Yu,~\IEEEmembership{Senior Member,~IEEE,}
        %and Yi Yu
        and Chia-Wen Lin,~\IEEEmembership{Fellow,~IEEE}
        % <-this % stops a space
\thanks{This work was supported in part by the foundation of Key Laboratory of Artificial Intelligence of Ministry of Education under Grant AI202404, and the Open Fund Project of Provincial Key Laboratory for Computer Information Processing Technology (Soochow University) under Grant KJS2274.~\textit{(Yangyang Qiu and Guoan Xu contributed equally to this work.)}~\textit{(Corresponding author: Guangwei Gao.)}}
\thanks{Yangyang Qiu and Guangwei Gao are with the Institute of Advanced Technology, Nanjing University of Posts and Telecommunications, Nanjing 210046, China, Key Laboratory of Artificial Intelligence, Ministry of Education, Shanghai 200240, and also with the Provincial Key Laboratory for Computer Information Processing Technology, Soochow University, Suzhou 215006, China (e-mail: csggao@gmail.com, 2513084587@qq.com).}% <-this % stops a space
\thanks{Guoan Xu is with the Faculty of Engineering and Information Technology, University of Technology Sydney, Sydney, Australia (e-mail: guoan.xu@student.uts.edu.au).}% <-this % stops a space
\thanks{Zhenhua Guo is with the Tianyijiaotong Technology Ltd., Suzhou 215100, China (e-mail: zhenhua.guo@tyjt-ai.com).}% <-this % stops a space
%\thanks{Huimin Lu is with the School of Automation, Southeast University, Nanjing 210096, China (e-mail: riku@cntl.Kyutech.ac.jp).}% <-this % stops a space
\thanks{Yi Yu is with the Graduate School of Advanced Science and Engineering, Hiroshima University, Hiroshima 739-8511, Japan (e-mail: yiyu@hiroshima-u.ac.jp).}% <-this % stops a space
\thanks{Chia-Wen Lin is with the Department of Electrical Engineering, National Tsing Hua University, Hsinchu, Taiwan 30013, R.O.C. (e-mail: cwlin@ee.nthu.edu.tw).}% <-this % stops a space
}

%\iffalse
\markboth{Journal of \LaTeX\ Class Files,~Vol-, No-, 2020}%
{Shell \MakeLowercase{\textit{et al.}}: Bare Demo of IEEEtran.cls for IEEE Journals}   
%\fi

\maketitle

\begin{abstract}
% In recent years, Convolutional Neural Networks (CNNs) and Transformers each have their own strengths in semantic segmentation tasks. However, the substantial computational burden and redundant parameters remain obstacles to the development of semantic segmentation. 
Recently, integrating the local modeling capabilities of Convolutional Neural Networks (CNNs) with the global dependency strengths of Transformers has created a sensation in the semantic segmentation community. However, substantial computational workloads and high hardware memory demands remain major obstacles to their further application in real-time scenarios. In this work, we propose a Lightweight Multiple-Information Interaction Network (LMIINet) for real-time semantic segmentation, which effectively combines CNNs and Transformers while reducing redundant computations and memory footprints. It features Lightweight Feature Interaction Bottleneck (LFIB) modules comprising efficient convolutions that enhance context integration. Additionally, improvements are made to the Flatten Transformer by enhancing local and global feature interaction to capture detailed semantic information. Incorporating a combination coefficient learning scheme in both LFIB and Transformer blocks facilitates improved feature interaction. Extensive experiments demonstrate that LMIINet excels in balancing accuracy and efficiency. With only 0.72M parameters and 11.74G FLOPs (Floating Point Operations Per Second), LMIINet achieves 72.0\% mIoU at 100 FPS (Frames Per Second) on the Cityscapes test set and 69.94\% mIoU (mean Intersection over Union) at 160 FPS on the CamVid test dataset using a single RTX2080Ti GPU.
\end{abstract}

% Note that keywords are not normally used for peer-reviewed papers.

% \fi
\IEEEpeerreviewmaketitle

\begin{figure}[t]
        \centering
	\includegraphics[width=0.5\textwidth]{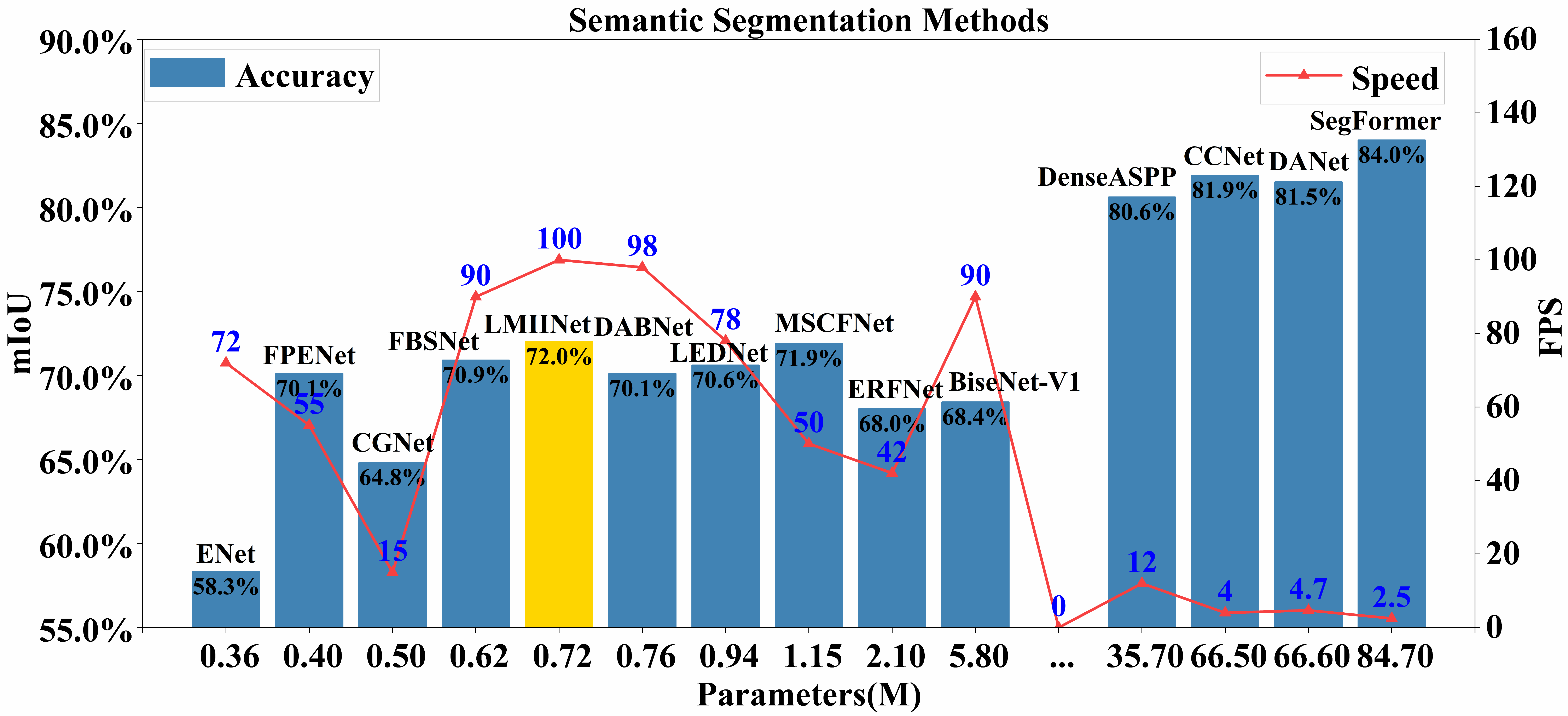}
	\caption{Accuracy-Parameters-Speed evaluations on the Cityscapes test dataset under the same device.}
	\vspace{-1.5em}
    \label{fig:blance}
\end{figure}

\section{INTRODUCTION}
\label{sec1}

\IEEEPARstart{S}{emantic} segmentation is a critical task in computer vision aimed at assigning each pixel in an image to specific semantic categories, enabling precise image segmentation~\cite{gong2023integrated,ding2022rethinking}. This technology finds applications in fields like autonomous driving and medical image analysis. Following the introduction of the Fully Convolutional Network (FCN)~\cite{long2015fully}, subsequent semantic segmentation models have evolved to focus on enhancing semantic information for dense tasks. However, FCN-based models face limitations in receptive field size. To address this challenge, alternative strategies include using large convolutional kernels~\cite{peng2017large}, dilated convolutions~\cite{chen2017deeplab}, and feature pyramids~\cite{zhao2017pyramid} to expand the receptive field. Additionally, attention mechanisms are utilized to integrate spatial and semantic information from different layers, enhancing feature map representational capacity.

The success of Transformers~\cite{vaswani2017attention} in natural language processing has led researchers to apply them in various downstream tasks, including Vision Transformer (ViT)~\cite{dosovitskiy2020image}, resulting in notable achievements. Transformers utilize self-attention mechanisms for global context modeling, enabling them to capture relationships within input sequences. This ability enhances semantic segmentation accuracy by improving the model's understanding of the image's semantic structure. Unlike traditional Convolutional Neural Networks (CNNs), Transformers excel at learning dependencies among distant pixels. However, Transformers primarily focuses on global context modeling, with limited emphasis on local spatial information in images. Additionally, the computational complexity of Transformers increases significantly with input sequence length, posing challenges in tasks like semantic segmentation with large image sizes. Therefore, striking a balance between global modeling strengths and computational efficiency is crucial when selecting a model for practical applications.

While Transformers excel in capturing global information, they exhibit limitations in extracting spatial details compared to CNNs. To address this, approaches integrating CNNs with Transformers are emerging to enhance semantic segmentation tasks. In the realm of semantic segmentation for mobile devices, methods like TopFormer~\cite{zhang2022topformer} have demonstrated promising results. Building on these advancements, we introduce LMIINet, a real-time semantic segmentation network based on a lightweight multiple-information interaction network. As depicted in Fig.~\ref{fig:blance}, LMIINet strikes a favorable balance among model performance, size, and inference speed. This paper makes contributions in three main aspects:
\begin{itemize}
    \item We introduce a Lightweight Feature Interaction Bottleneck (LFIB) module as a core component for feature extraction. LFIB integrates depth-wise separable convolution, asymmetric convolution, and dilated convolution techniques, significantly reducing the computational load. Additionally, LFIB incorporates combination coefficient learning, promoting efficient interaction of features.
    
    \item We introduce a hybrid network named LMIINet, tailored for semantic segmentation tasks. LMIINet utilizes an encoder-decoder architecture and integrates an enhanced Flatten Transformer to capture detailed semantic information. Additionally, LMIINet combines boundary detail information for context information fusion during the resolution recovery process.
    
    \item LMIINet achieved a mIoU of 72.0\% on the Cityscapes test set using the RTX2080Ti hardware platform with just 0.72M parameters. It also attained a strong performance of 69.94\% on the CamVid dataset, surpassing many existing models in terms of performance.
\end{itemize}

\begin{figure*}[!ht]
	\centering
	\includegraphics[width=0.97\textwidth]{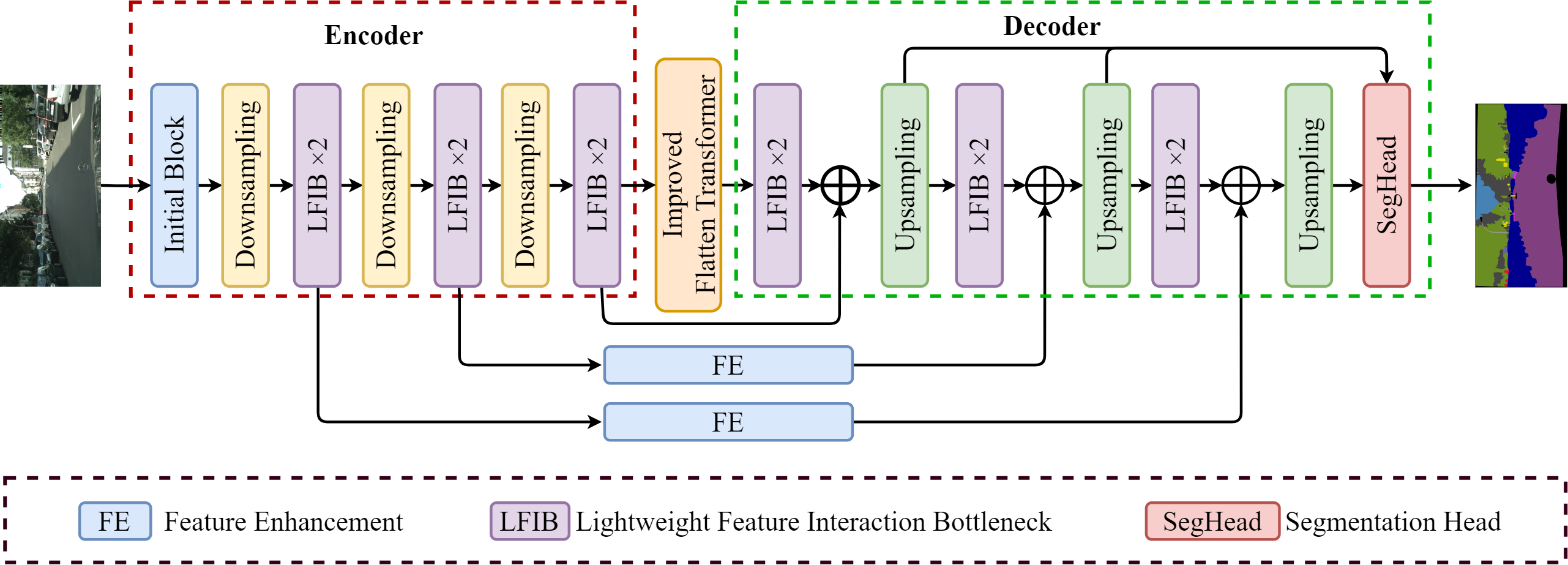}
	\caption{The complete architecture of the proposed Lightweight Multiple-Information Interaction Network (LMIINet). It consists of three parts: the decoding stage, the encoding stage, and the improved flatten Transformer.} 
	\label{fig:network}
\end{figure*}

\begin{figure*}[!ht]
        \centering
	\includegraphics[width=0.97\textwidth]{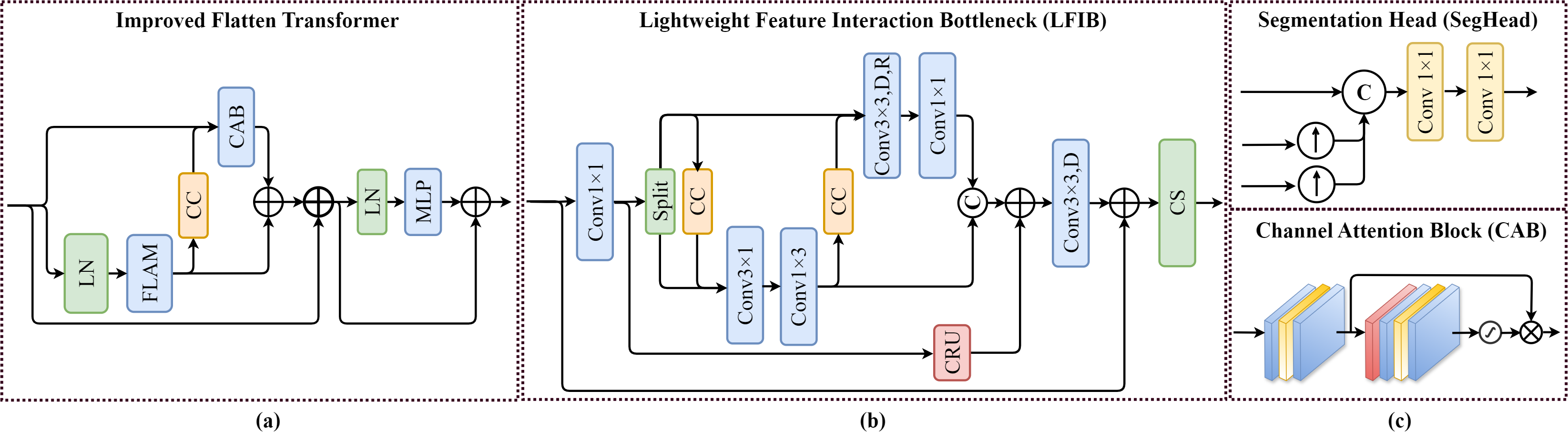}
	\caption{The diagram of the proposed Lightweight Feature Interaction Bottleneck (LFIB), improved Flatten Transformer, Segmentation Head (SegHead), and Channel Attention Block (CAB). $D$ represents the depth-wise convolution, $R$ is the kernel of dilated convolution, and $CS$ denotes the channel shuffle operation.}
	\vspace{-1.5em}
    \label{fig:block}
\end{figure*}

\section{RELATED WORK}
\label{sec2}

\subsection{CNN-based Semantic Segmentation Methods}
\label{sec21}

Significant advancements in semantic segmentation methods have been driven by the robust feature representation capabilities of Convolutional Neural Networks (CNNs). %The Fully Convolutional Network (FCN)~\cite{long2015fully} has been a seminal contribution in this field, serving as the basis for subsequent refined models. PSPNet~\cite{zhao2017pyramid} introduced the pyramid pooling module, pooling regions of varying scales in the input image to capture contextual information and enhance segmentation performance. 
BiSeNet~\cite{yu2018bisenet} and BiSeNet-v2~\cite{yu2021bisenet} proposed a dual-path network structure, extracting deep semantic and shallow spatial information to reduce computational costs without compromising performance. %Incorporating attention mechanisms enhances the model's understanding of global image relationships, crucial for tasks like semantic segmentation requiring long-distance context processing. Dynamic weight adjustments through attention mechanisms improve feature significance adaptability, enhancing segmentation accuracy and consistency. 
DeepLabV3+~\cite{chen2018encoder} introduced self-attention mechanisms to enhance contextual information capture in images, addressing challenges like object boundary ambiguity and multi-scale targets. DANet~\cite{fu2019dual} integrated attention mechanisms in spatial and channel dimensions, enabling comprehensive capture of long-distance dependencies in deep features, significantly boosting segmentation accuracy. AGLN~\cite{li2022attention} intricately designed a context fusion block to integrate global contexts from high-level features, refining encoder features in channel and spatial dimensions to enhance segmentation results.

While these approaches have shown success in enhancing results, they have not altered the fundamental nature of non-local algorithms as pixel-wise matrix operations, posing significant computational challenges. Consequently, lightweight networks have emerged as a solution. For example, ERFNet~\cite{romera2017erfnet} is a real-time semantic segmentation network that efficiently utilizes factorized convolutions and global average pooling. SwiftNet~\cite{wang2021swiftnet} employed lightweight blocks to reduce computational complexity in semantic segmentation tasks through the incorporation of SwiftBlock. ESPNet~\cite{mehta2018espnet} and ESPNet-v2~\cite{mehta2019espnetv2} decrease parameter count and computational load by decomposing convolutions into pointwise and dilated convolutions. NDNet~\cite{yang2020ndnet} analyzed the computational cost of CNNs and designed a narrow yet deep backbone network to enhance semantic segmentation efficiency. MSCFNet~\cite{gao2021mscfnet} introduced an efficient asymmetric residual module and a lightweight network using a multi-scale context fusion scheme with an asymmetric encoder-decoder architecture. MLFNet~\cite{fan2022mlfnet} constructed a lightweight backbone with an increased receptive field to encode pixel-level features, along with a spatial compensation branch to refine feature maps and enhance low-level details. FBSNet~\cite{gao2022fbsnet} utilized a symmetrical encoder-decoder structure with two branches: a semantic information branch for contextual information acquisition and a spatial detail branch to preserve local pixel dependencies and details.

\subsection{Transformer-based Semantic Segmentation Methods}
\label{sec22}

Originally designed for natural language processing tasks, Transformers have increasingly been applied in computer vision, including tasks such as semantic segmentation~\cite{xu2023lightweight}. %For instance, ViT~\cite{dosovitskiy2020image} was among the pioneering models to introduce the Transformer architecture to image processing. ViT segments images into one-dimensional sequences, utilizing the Transformer's self-attention mechanism for global modeling. It showcased impressive performance in image classification, spurring further exploration of Transformers in semantic segmentation. 
SETR~\cite{zheng2021rethinking} is a Transformer model tailored for semantic segmentation, featuring both local attention layers and global self-attention layers to capture multi-scale image information. %SETR's architecture enables it to operate at arbitrary resolutions, enhancing adaptability to images of varying sizes. TransUNet~\cite{chen2021transunet} integrates Transformer and UNet for medical image segmentation, replacing the traditional CNN encoder with a Transformer encoder to improve global context modeling. 
SegFormer~\cite{xie2021segformer} is a semantic segmentation model built on Transformer's encoder-decoder structure, introducing an interaction mechanism for local and global features, showcasing strong performance on diverse datasets. TopFormer~\cite{zhang2022topformer} is a mobile-friendly architecture that incorporates tokens from different scales to generate scale-aware semantic features, enhancing representation. LMFFNet~\cite{shi2022lmffnet} designed a lightweight multiscale-feature-fusion network and achieved a good balance between parameters and accuracy. AFFormer~\cite{dong2023head} adopted a parallel architecture utilizing prototype representations as specific local descriptors, replacing the decoder to preserve rich image semantics in high-resolution features. It introduces a lightweight prototype learning block with linear complexity to substitute standard self-attention mechanisms.

Although the methods discussed above have shown promising results, the Transformer's limited spatial information tends to produce coarse segmentation results. In contrast, CNNs excel in addressing spatial information deficiencies. Therefore, our objective is to explore an approach that integrates multiple-information interactions between CNNs and Transformers to enhance the efficiency of segmentation tasks. Existing methods have limitations in lightweight design and global semantic modeling, which are the key focus areas for improvement in the proposed model.

\iffalse
\begin{figure}[t]
        \centering
	\includegraphics[width=0.5\textwidth]{figs/3.png}
	\caption{The proposed Lightweight Feature Interaction Bottleneck (LFIB). Among them, $D$ represents depth-wise convolution, $R$ is the kernel of dilated convolution, and $CS$ denotes the channel shuffle operation.}
	\label{fig:LFIB}
\end{figure}
\fi

\begin{figure*}[t]
        \centering
	\includegraphics[width=0.97\textwidth]{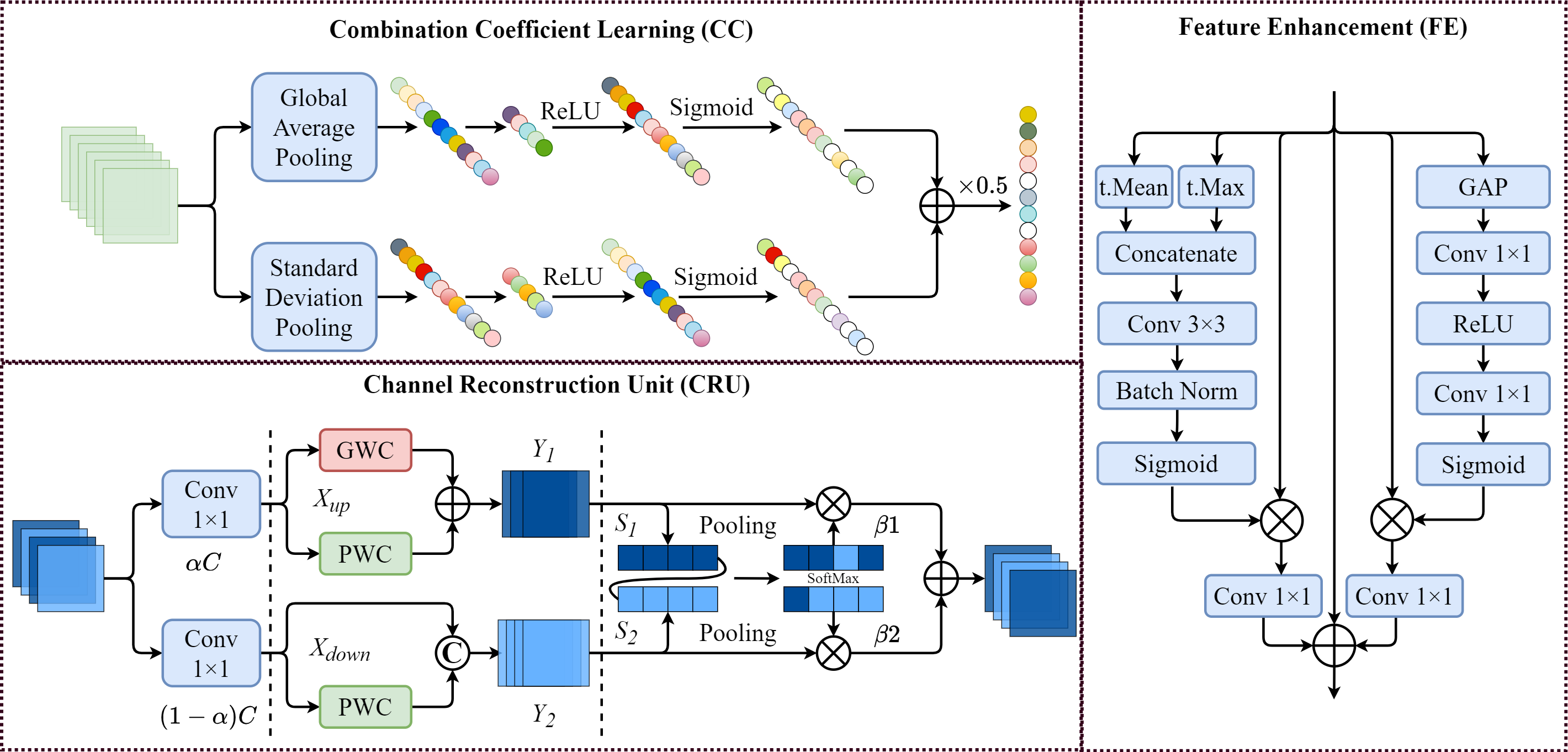}
	\caption{The diagram of the Combination Coefficient learning (CC) scheme, Channel Recurrent Unit (CRU), and the Feature Enhancement (FE) module.}
	\vspace{-1.5em}
    \label{fig:auxblock}
\end{figure*}

\section{PROPOSED METHOD}
\label{sec3}

\subsection{Network Architecture}
\label{sec31}

As depicted in Fig.~\ref{fig:network}, the architecture of LMIINet shares similarities with UNet, consisting of an encoder, a decoder, an improved Flatten Transformer, and two long connection modules. The encoder and decoder in LMIINet utilize customized CNN structures to extract local feature information, enhancing feature representation efficacy. Additionally, a variation of the Flatten Transformer~\cite{han2023flatten} is integrated between the encoder and decoder to focus on modeling global feature dependencies. Inspired by LETNet~\cite{xu2023lightweight}, the long connection modules serve as an adaptive information transfer mechanism that bridges the encoder-decoder gap. By complementing the decoder with absent object details, it aids in resolution recovery, ultimately improving segmentation accuracy.

\subsection{Lightweight Feature Interaction Bottleneck (LFIB)}
\label{sec32}

As illustrated in Fig.~\ref{fig:block} (b), the LFIB structure is designed based on the residual concept to achieve feature extraction with fewer network layers. The LFIB module is designed to reduce computational costs through asymmetric convolution and depthwise separable convolution while enhancing the ability to interact with multi-scale features. At the bottleneck, a $1\times1$ convolution is applied to maintain the same number of feature channels, ensuring feature consistency. Subsequently, the output is divided into two parts with equal channel numbers, each entering a dual-branch structure. The left branch employs a combination of $3\times1$ and $1\times3$ asymmetric convolutions to enhance the network's representational capacity without increasing parameters, offering flexibility in adjusting the receptive field size. The right branch consists of $3\times3$ Depth-Wise (DW) and Point-Wise (PW) convolutions. DW convolution conducts independent convolutions on each input channel, reducing parameters and capturing local spatial information effectively. PW convolution integrates information between channels post-DW convolution. To promote feature interaction, the Combination Coefficient learning (CC) scheme (depicted in Fig.~\ref{fig:auxblock} focuses on the weighted processing of feature maps from different channels, highlighting essential features. The CC is positioned in the dual-branch structure, weighting the left branch output and incorporating it into the right branch input. Experimental findings demonstrate that CC significantly enhances segmentation performance. The above operations are denoted as follows:
\begin{equation}
   \begin{aligned}
&F_{1}=H_{1\,\times\,1}\,({x_l}), \\
   \end{aligned}
\end{equation}
\begin{equation}
   \begin{aligned}
&F_{21}, F_{22}=H_{Split}\,(F_{1}), \\
   \end{aligned}
\end{equation}
\begin{equation}
   \begin{aligned}
&F_{21}=H_{1\,\times\,3}\,(H_{3\,\times\,1}\,(F_{21}+H_{CC}\,(F_{22}))), \\
   \end{aligned}
\end{equation}
\begin{equation}
   \begin{aligned}
&F_{22}=H_{1\,\times\,1}(H_{3\,\times\,3,D,R}\,(F_{22}+H_{CC}\,(F_{21}))), \\
   \end{aligned}
\end{equation}
where ${x_l}$ signifies the input feature maps of LFIB, $H_{k \times k}\left( \cdot \right)$ denotes the convolution operation, $H_{Split}$ indicates the channel split operation, and $H_{CC}$ denotes the CC learning scheme.

The outputs from the dual branches are concatenated, and the intermediate features are processed through the Channel Recurrent Unit (CRU) operation~\cite{li2023scconv}, which enriches representative features using lightweight convolution operations. CRU efficiently manages redundant features through a cost-effective operation and feature reuse scheme. The CRU structure is illustrated in Fig.~\ref{fig:auxblock}. Subsequently, the output undergoes a $3\times3$ depth-wise convolution for further feature extraction. To address issues related to information independence and lack of correlation between channels introduced by deep convolution, a channel shuffle strategy is implemented to promote semantic information exchange among different channels. These aforementioned operations are shown as follows:
\begin{equation}
   \begin{aligned}
&F_{2}={Conact\,\left({F_{21}},{F_{22}} \right)} + H_{CRU}\,(F_{1}), \\
   \end{aligned}
\end{equation}
\begin{equation}
   \begin{aligned}
&{y_l}=H_{CS}\,(H_{3\,\times\,3,D}\,({x_l}+F_{2})), \\
   \end{aligned} 
\end{equation}
where $y_l$ represents the output feature maps of LFIB, $H_{CRU}$ represents the CRU operation, $Conact\left( \cdot \right)$ denotes the concatenation operation along the channel dimension, and $H_{CS}$ indicates the channel shuffle operation.

\iffalse
\begin{figure}[t]
        \centering
	\includegraphics[width=0.5\textwidth]{figs/4.png}
    \hfil
	\caption{Schematic diagram of the (a) Channel Reconstruction Unit (CRU) and (b) Combination Coefficient Learning ($CC$). Please zoom in for details.}
	\label{fig:4}
\end{figure}

\begin{figure}[t]
\centering
	\includegraphics[width=0.45\textwidth]{figs/5.png}
    \hfil
	\caption{Schematic diagram of the (a) Improved Flatten Transformer and
 (b) Feature Enhancement (FE) module. Please zoom in for details.}
	\label{fig:5}
\end{figure}

\begin{figure}[t]
\centering
	\includegraphics[width=0.45\textwidth]{figs/6.png}
    \hfil
	\caption{Schematic diagram of (a) the SegHead and the (b) 
 Channel Attention Block($CAB$). Please zoom in for details.}
	\label{fig:6}
\end{figure}
\fi

\subsection{Improved Flatten Transformer}
\label{sec33}

While CNNs are proficient in extracting local features, their ability to capture comprehensive global semantic information is limited. To address this, we leverage the Transformer to facilitate the learning of global features. However, existing methods such as ViT~\cite{dosovitskiy2020image} and Swin Transformer~\cite{liu2021swin} do not yield optimal performance in semantic segmentation tasks. Therefore, we choose to implement the Flatten Transformer~\cite{han2023flatten}, which introduces a novel Focused Linear Attention Module (FLAM) replacing the traditional self-attention module in Transformers. This enhancement aims to enhance efficiency and expressiveness in semantic segmentation tasks.

The Flatten Transformer introduces a novel linear attention module called the FLAM, to reduce computational complexity while maintaining expressive capabilities. Initially, a novel mapping function is crafted to replicate the focused distribution of the original $Softmax$ attention mechanism. Additionally, to overcome the low-rank limitation observed in prior linear attention modules, a straightforward depth-wise convolution is utilized to reintroduce feature diversity. This strategy allows the new module to combine the advantages of linear complexity and enhanced expressiveness simultaneously. In particular, the FLAM module can be formulated as
\begin{equation}
   \begin{aligned}
&y_f=Sim(Q,K)V=\phi_{P}({Q})\,\phi_{P}({K})^{T}V+H_{DWC}({V}),\\
   \end{aligned}
\end{equation}
\begin{equation}
   \begin{aligned}
&\phi_{P}({x_f})=H_{P}\,(ReLU(x_f)),\;H_{P}\,(x_f)=(\frac{\Vert {x_f} \Vert}{\Vert {x_f}^{P} \Vert}){x_f}^{P}, \\
   \end{aligned}
\end{equation}
where $x_f$ and $y_f$ represent the input and output of FLAM, $Q$, $K$, and $V$ represent the query, key, and value metrices, $H_{p}$ denotes the focused function, $H_{DWC}$ means the depth-wise convolution operation.

To enhance the capture of detailed semantic information, we implemented specific adjustments to the Flatten Transformer, as shown in Fig.~\ref{fig:block} (a). Within the dual-branch structure at the FLAM position, we integrated a Channel Attention Block (CAB) module~\cite{chen2023activating}. Research suggests that utilizing channel attention can activate more pixels by incorporating global information in channel attention weight computation. This integration can improve the Transformer's visual representations and aid optimization. Therefore, we incorporated a convolution block based on channel attention into the standard Transformer block to boost the network's representational capacity. The CAB module comprises two standard convolution layers, one with GELU activation~\cite{hendrycks2016gaussian}, and the other involving a CAB, as depicted in Fig.~\ref{fig:block} (c). Additionally, following the LFIB approach, we applied the CC scheme for information interaction within the dual-branch structure. Experimental findings suggest that these modifications enhance performance. All of these steps can be summarized as follows:
\begin{equation}
   \begin{aligned}
&F_{11}=H_{FLAM}\,(H_{LN}\,({x_t})), \\
   \end{aligned}
\end{equation}
\begin{equation}
   \begin{aligned}
&F_{12}=H_{CAB}\,(x_t+H_{CC}\,(F_{11})), \\
   \end{aligned}
\end{equation}
\begin{equation}
   \begin{aligned}
&F_{1}=F_{11}+F_{12}+{x_t}, \\
   \end{aligned}
\end{equation}
\begin{equation}
   \begin{aligned}
&{y_t}=F_{1}+H_{MLP}\,(H_{LN}\,(F_{1})), \\
   \end{aligned} 
\end{equation}
where ${x_t}$ and ${y_t}$ represent the input and output of the Flatten Transformer, $H_{FALM}$ refers to the FLAM module, $H_{CAB}$ indicates the CAB operation, $H_{LN}$ denotes the layer-normalization operation, and $H_{MLP}$ signifies the feed-forward operation.

\subsection{Others}
\label{sec34}

Moreover, we drew inspiration from the Feature Enhancement (FE) module in LETNet~\cite{xu2023lightweight}, illustrated in Fig.~\ref{fig:auxblock}. This module is employed to merge low-level spatial information with high-level semantic information. In segmentation tasks, it is essential to seamlessly integrate low-level spatial details with high-level semantic content to ensure that high-level information retains adequate spatial context, thereby facilitating high-quality image segmentation.

After incorporating semantic information, features obtained at various scales contain a blend of detailed spatial and semantic information, which is vital for semantic segmentation tasks. Our used segmentation head (SegHead) implements an integration approach by amalgamating features that have been downsampled to 1/2, 1/4, and 1/8, as depicted in Fig.~\ref{fig:block} (c). Initially, we upsample the lower-resolution features to align with the dimensions of the higher-resolution features. Subsequently, through concatenation, we fuse features from all scales. Finally, following two convolution layers, the final segmentation map is produced. All of these steps can be summarized as follows:
\begin{equation}
   \begin{aligned}
&F = {\sum\limits_{i = 1}^n H _{unsample}}({x_{s_i}}), \\
   \end{aligned}
\end{equation}

\begin{equation}
   \begin{aligned}
&y_{s} = {H_{1\,\times\,1}}({H_{1\,  \times\,1}}(F)), \\
   \end{aligned}
\end{equation}
where ${x_{s_i}}$ and $y_{s}$ represent the input and output of the segmentation head, $H_{unsample}$ denotes the upsampling operation.

\section{EXPERIMENTS}
\label{sec4}

\subsection{Datasets}
\label{sec41}

\textbf{Cityscapes~\cite{cordts2016cityscapes}:} The dataset comprises images with a resolution of $2048 \times 1024$, sourced from urban road scenes in 50 cities in Germany and France, featuring elements such as pedestrians, roads, and vehicles. It encompasses 19 categories for semantic segmentation evaluation. Out of the 5000 meticulously annotated images, 2075 are allocated for training, 500 for validation, and 1525 for testing purposes.

\textbf{Camvid~\cite{brostow2008segmentation}:} The dataset comprises 701 urban road images sized $960 \times 720$, categorized into 11 classes. Among the finely annotated images, 367 are allocated for training, 101 for validation, and 233 for testing.

\begin{table}[t]
	\setlength{\abovecaptionskip}{0pt}
	\setlength{\belowcaptionskip}{10pt}
	\caption{The details of the model settings.}
	\label{tab:modelsettings}
        \begin{center}
        \renewcommand\arraystretch{1.3}
	\begin{tabular}{ccc}
\toprule
			Dataset & Cityscapes & CamVid \\
\midrule
    Batch Size & 4 & 8 \\

    Loss function & \multicolumn{2}{c}{CrossEntropy Loss}  \\

    Optimization method & SGD(momentum 0.9) & Adam(momentum 0.9) \\

    Weight decay & ${1}\times{10}^{-4}$ & ${2}\times{10}^{-4}$  \\

    Initial learning rate & ${4.5}\times{10}^{-2}$ & ${1}\times{10}^{-3}$ \\
                
    Learning rate policy & \multicolumn{2}{c}{Poly} \\
\bottomrule
		\end{tabular}
            \end{center}
\end{table}

\begin{table}[t]
	\setlength{\abovecaptionskip}{0pt}
	\setlength{\belowcaptionskip}{10pt}
	\caption{Ablation study of the improved Flatten Transformer. $\ast$ represents the final version. Baseline represents the model without Flatten Transformer}
	\label{tab:Transformer}
        \begin{center}
        \renewcommand\arraystretch{1.1}
	\begin{tabular}{p{2.4cm}p{1.1cm}<{\centering}p{1.1cm}<{\centering}p{1.1cm}<{\centering}p{1.1cm}<{\centering}}
\toprule
	Model & Parameter (K)$\downarrow$ & FLOPs (G)$\downarrow$ &  Speed (FPS)$\uparrow$ & mIoU ($\%$)$\uparrow$ \\
\midrule
$Baseline$ & \colorbox[RGB]{236,236,236}{654K} & \colorbox[RGB]{215,215,215}{3.54} & 170 & 68.79   \\

$+ Trans$  & 698K & 3.56 & \colorbox[RGB]{215,215,215}{190} & $69.01^{ \color{blue}{0.22\uparrow}}$   \\

$+ Trans + CAB$ & 718K & 3.63 & 180 & $69.18^{ \color{blue}{0.39\uparrow}}$  \\

$+ Trans + CAB + CC^\ast$ & 720K & 3.63 & 160 & \colorbox[RGB]{194,194,194}{$69.94^{ \color{blue}{1.15\uparrow}}$}   \\
        
\bottomrule
	\end{tabular}
        \end{center}
\end{table}

\begin{table}[t]
	\setlength{\abovecaptionskip}{0pt}
	\setlength{\belowcaptionskip}{10pt}
	\caption{Ablation study of the CC scheme. \Checkmark represents with $CC$.}
	\label{tab:cc}
        \begin{center}
        \renewcommand\arraystretch{1.1}
	\begin{tabular}{p{0.5cm}<{\centering}p{0.5cm}<{\centering}p{1.3cm}<{\centering}p{1.3cm}<{\centering}p{1.3cm}<{\centering}p{1.3cm}<{\centering}}
\toprule
		LFIB & Trans & Param. (K)$\downarrow$ & FLOPs (G)$\downarrow$ & Speed (FPS)$\uparrow$ & mIoU ($\%$)$\uparrow$ \\
\midrule

\XSolidBrush & \XSolidBrush & \colorbox[RGB]{236,236,236}{691} & \colorbox[RGB]{236,236,236}{3.62} & \colorbox[RGB]{215,215,215}{200} & 68.60   \\

\Checkmark & \XSolidBrush & 718 & 3.63 & 180 & $69.18^{ \color{blue}{0.58\uparrow}}$  \\

\XSolidBrush & \Checkmark & 694 & \colorbox[RGB]{236,236,236}{3.62} & 190 & $69.13^{ \color{blue}{0.53\uparrow}}$  \\

\Checkmark & \Checkmark & 720 & 3.63 & 160 & \colorbox[RGB]{194,194,194}{$69.94^{ \color{blue}{1.34\uparrow}}$}  \\ 

\bottomrule
	\end{tabular}
        \end{center}
\end{table}

\begin{table}[t]
	\setlength{\abovecaptionskip}{0pt}
	\setlength{\belowcaptionskip}{10pt}
	\caption{Ablation study of the SegHead. $\ast$ represents the final version.}
	\label{tab:seghead}
        \begin{center}
        \renewcommand\arraystretch{1.1}
	\begin{tabular}{p{2.4cm}p{1.0cm}<{\centering}p{1.0cm}<{\centering}p{1.0cm}<{\centering}p{1.0cm}<{\centering}}
\toprule
        Methods & Param. (K)$\downarrow$ & FLOPs (G)$\downarrow$  & Speed (FPS)$\uparrow$ & mIoU ($\%$)$\uparrow$ \\
\midrule

$\left\{1/2\right\}$ & \colorbox[RGB]{236,236,236}{713} & \colorbox[RGB]{215,215,215}{3.31} & \colorbox[RGB]{215,215,215}{170} & 69.18 \\

$\left\{1/2,1/4\right\}$ & 716 & 3.45 & 130 & $69.20^{ \color{blue}{0.02\uparrow}}$ \\

$\left\{1/2,1/4,1/8\right\}^\ast$ & 720 & 3.63 & 160 & \colorbox[RGB]{194,194,194}{$69.94^{ \color{blue}{0.76\uparrow}}$} \\

$\left\{1/2,1/4,1/8,1/16\right\}$ & 722 & 3.72 & 150 & $69.47^{ \color{blue}{0.29\uparrow}}$ \\
\bottomrule
      \end{tabular}
      \end{center}
\end{table}

\begin{table}[t]
	\setlength{\abovecaptionskip}{0pt}
	\setlength{\belowcaptionskip}{10pt}
	\caption{Ablation study of the network layers on our model. $\ast$ represents the final version.}
	\label{tab:nofLFIB}
        \begin{center}
        \renewcommand\arraystretch{1.1}
	\begin{tabular}{p{1.8cm}p{1.2cm}<{\centering}p{1.2cm}<{\centering}p{1.2cm}<{\centering}p{1.2cm}<{\centering}}
\toprule
     Layers & Param. (K)$\downarrow$ & FLOPs (G)$\downarrow$  & Speed (FPS)$\uparrow$  & mIoU ($\%$)$\uparrow$ \\
\midrule
   $\left\{1,1,6,6,1,1\right\}$ & \colorbox[RGB]{236,236,236}{704}  & \colorbox[RGB]{215,215,215}{3.23} & 120 & 67.83   \\
   
   $\left\{1,2,4,4,2,1\right\}$ & 745 & 3.42 & 140 & $68.17^{ \color{blue}{0.34\uparrow}}$   \\
   
    $\left\{1,2,3,3,2,1\right\}$ & 720 & 3.40 & 130 & $69.51^{ \color{blue}{1.68\uparrow}}$  \\
			
    $\left\{2,2,2,2,2,2\right\}^\ast$ & 720 & 3.63 & \colorbox[RGB]{215,215,215}{160} & \colorbox[RGB]{194,194,194}{$69.94^{ \color{blue}{2.11\uparrow}}$}  \\
\bottomrule
    \end{tabular}
    \end{center}
\end{table}

\begin{table}[t]
	\setlength{\abovecaptionskip}{0pt}
	\setlength{\belowcaptionskip}{10pt}
	\caption{Ablation study of the auxiliary losses on our model. $\ast$ represents the final version. The baseline represents the version without the auxiliary loss.}
	\label{tab:Auxloss}
        \begin{center}
        \renewcommand\arraystretch{1.1}
	\begin{tabular}{p{1.3cm}p{1.3cm}<{\centering}p{1.3cm}<{\centering}p{1.3cm}<{\centering}p{1.3cm}<{\centering}}
\toprule
		Methods & Param. (K)$\downarrow$ & FLOPs (G)$\downarrow$ & Speed (FPS)$\uparrow$ & mIoU ($\%$)$\uparrow$ \\
\midrule

   Baseline & \colorbox[RGB]{236,236,236}{719} & \colorbox[RGB]{215,215,215}{3.38} & \colorbox[RGB]{215,215,215}{190} & 68.58 \\

   1/2 & \colorbox[RGB]{236,236,236}{719} & 3.45 & 150 & $68.71^{ \color{blue}{0.13\uparrow}}$ \\

   1/4 & 720 & 3.51 & 150 & $69.71^{ \color{blue}{1.13\uparrow}}$ \\

    $1/8^\ast$ & 720 & 3.63 & 160 & \colorbox[RGB]{194,194,194}{$69.94^{ \color{blue}{1.36\uparrow}}$} \\

    1/16 & 720 & 3.51 & 130 & $68.63^{ \color{blue}{0.05\uparrow}}$ \\
    
\bottomrule
		\end{tabular}
            \end{center}
\end{table}

\subsection{Model Settings}
\label{sec42}

In this study, we utilize the PyTorch framework to construct the model and conduct training on an RTX2080Ti GPU. Table~\ref{tab:modelsettings} presents the model settings of LMIINet for the Cityscapes and CamVid datasets. The batch size for Cityscapes and CamVid was set to 4 and 8, respectively, to accommodate differences in dataset characteristics and GPU memory limitations. Cityscapes contains higher-resolution images, which demand more memory, thus requiring a smaller batch size. In contrast, CamVid images have lower resolution, enabling a slightly larger batch size. The learning rate is dynamic across iterations and can be computed as follows:
\begin{equation}
\begin{aligned}
&lr = {lr_{initial} \times (1 - {\frac{iteration}{max\_iteration})^{0.9}}},\\
\end{aligned}
\end{equation}
where ${lr_{initial}}$ represents the initial learning rate. It is important to highlight that we train Cityscapes and CamVid datasets separately with distinct parameter configurations due to variations in dataset resolution. Alongside the primary loss function ($loss_{main}$) monitoring the network's overall output, we have incorporated an auxiliary loss function ($loss_{aux}$) to improve model training supervision. The formula for the loss function can be expressed as
\begin{equation}
\begin{aligned}
&loss={loss_{main} + { \lambda \times loss_{aux}}}.\\
\end{aligned}
\end{equation}
The loss function $loss_{main}$ compares the segmentation head's output to the ground truth, while $loss_{aux}$ measures the discrepancy between the output of a specific stage in the decoding process post-direct upsampling and the ground truth. The parameter $\lambda$ denotes the weight assigned to the auxiliary loss function. Experimental findings show that the segmentation accuracy reaches its highest point when $\lambda$ is set to ${0.3}$.

\begin{table*}[t]
	\setlength{\abovecaptionskip}{2pt}
	\setlength{\belowcaptionskip}{10pt}
	\caption{Comparisons with the state-of-art semantic segmentation methods on the Cityscapes test dataset.}
	\label{tab:city}
        \centering
        \setlength\tabcolsep{7.2pt}
        \renewcommand\arraystretch{1.1}
	\begin{tabular}
 {clcccccccc}
 %{p{0.1cm}p{2cm}p{0.5cm}<{\centering}p{1.4cm}<{\centering}p{1.4cm}<{\centering}p{1.2cm}<{\centering}p{1.0cm}<{\centering}p{2.0cm}<{\centering}p{1.0cm}<{\centering}p{1.0cm}<{\centering}}
\toprule
     & Methods & Year & Resolution & Backbone & Param. (M)$\downarrow$  & FLOPs (G)$\downarrow$ & GPU & Speed (FPS)$\uparrow$ & mIoU ($\%$)$\uparrow$ \\
\midrule

      \multirow{7}{*}{\rotatebox{90}{Large Size}} %& DeepLab~\cite{chen2017deeplab} & 2015 & $512 \times 1024$ & ResNet-101 & 262.10 & 457.8 & 0.25 & 63.5\\
      
      & DeepLabV3+~\cite{chen2018encoder} & 2018 & $1024 \times 2048$ & Xception & \colorbox[RGB]{236,236,236}{15.40} & \colorbox[RGB]{215,215,215}{555.4} & Titan X & 8.4 & $75.2^{ \color{red}{3.2\downarrow}}$ \\
 
      & DenseASPP~\cite{yang2018denseaspp} & 2018 & $512 \times 512$ & DenseNet & 35.70 & 632.9 & Titan XP & \colorbox[RGB]{194,194,194}{12} & 
      $80.6^{ \color{red}{8.6\downarrow}}$\\

      & DANet~\cite{fu2019dual} & 2019 & $1024 \times 1024$ & ResNet-101 & 66.60 & 1298.8 & Titan XP & 4 & 
      $81.5^{ \color{red}{9.5\downarrow}}$\\

     & CCNet~\cite{huang2019ccnet} & 2019 & $1024 \times 1024$ & ResNet-101 & 66.50 & 1153.9 & Titan XP & 4.7 & 
     $81.9^{ \color{red}{9.9\downarrow}}$\\

    & SETR~\cite{zheng2021rethinking} & 2021 & $768 \times 768$ & ViT-Large & 318.30 & - & 1080Ti & 0.5 & 
    $82.2^{ \color{red}{10.2\downarrow}}$\\

    & SegFormer~\cite{xie2021segformer} & 2021 & $1024 \times 2048$ & MiT-B5 & 84.70 & 1447.6 & 1080Ti & 2.5 & \colorbox[RGB]{183,183,183}{$84.0^{ \color{red}{12.0\downarrow}}$} \\
\midrule

        \multirow{8}{*}{\rotatebox{90}{Medium Size}} %& SegNet~\cite{badrinarayanan2017segnet} & 2017 & $720 \times 1280$ & VGG-16 & 29.50 & 286.0 & Titan X & 3.5 & $57.0^{ \color{red}{5\downarrow}}$ \\

     & ERFNet~\cite{romera2017erfnet} & 2017 & $512 \times 1024$ & No & \colorbox[RGB]{236,236,236}{2.10} & - & Titan X & 42 & $68.0^{ \color{blue}{4.0\uparrow}}$ \\
 
      & BiseNet-v1~\cite{yu2018bisenet} & 2018 & $768 \times 1536$ & Xception & 5.80 & 14.8 &  Titan XP & 90 & 
      $68.4^{ \color{blue}{3.6\uparrow}}$\\
        
      & ICNet~\cite{zhao2018icnet} & 2018 & $1024 \times 2048$ & PSPNet-50 & 26.50 & 28.3 & Titan X & 30 & 
      $69.5^{ \color{blue}{2.5\uparrow}}$\\

      & DFANet~\cite{li2019dfanet} & 2019 & $1024 \times 1024$ & Xeption & 7.80 & \colorbox[RGB]{215,215,215}{3.4} & Titan X & \colorbox[RGB]{194,194,194}{100} & $71.3^{ \color{blue}{0.7\uparrow}}$ \\

     & STDC1-50~\cite{fan2021rethinking} & 2021 & $512 \times 1024$ & - & 8.40 & - & 1080Ti & 87 & $71.9^{ \color{blue}{0.1\uparrow}}$ \\

    & MLFNet~\cite{fan2022mlfnet} &2023 & $512 \times 1024$ & ResNet-34      & 13.03     & 15.5 & Titan XP  & 72  & 
    $72.1^{ \color{red}{0.1\downarrow}}$\\ 

    & BiseNet-v2~\cite{yu2021bisenet} & 2021 & $512 \times 1024$ & Xception & 3.40 & 21.2 & 1080Ti & 92 & \colorbox[RGB]{183,183,183}{$72.6^{ \color{red}{0.6\downarrow}}$} \\

\midrule

     \multirow{20}{*}{\rotatebox{90}{Small Size}} %& ENet~\cite{paszke2016enet} & 2016 & $720 \times 1280$ & No & \colorbox[RGB]{236,236,236}{0.36} & 3.8 & Titan X & 46.8 & $58.3^{ \color{red}{13.7\downarrow}}$\\

    & ESPNet~\cite{mehta2018espnet} & 2018 & $512 \times 1024$ & No & \colorbox[RGB]{236,236,236}{0.36} & - & Titan XP & 93 & $60.3^{ \color{blue}{11.7\uparrow}}$ \\

    & CGNet~\cite{wu2020cgnet} & 2020 & $360 \times 640$ & No & 0.50 & 6.0 & V100 & 17.6 & $64.8^{ \color{blue}{7.2\uparrow}}$ \\

    & NDNet~\cite{yang2020ndnet} & 2021 & $1024 \times 2048$ & No & 0.50 & 14.0 & 2080Ti & 40 & $65.3^{ \color{blue}{6.7\uparrow}}$ \\

    & ESPNet-v2~\cite{mehta2019espnetv2} & 2019 & $512 \times 1024$ & No & 3.49 & \colorbox[RGB]{215,215,215}{2.7} & Titan Xp & 80 & $66.2^{ \color{blue}{5.8\uparrow}}$ \\

    & SCMNet~\cite{singha2021scmnet} & 2021 & $512 \times 1024$ & No & 1.20 & 18.2& 2080Ti & 92 & $68.3^{ \color{blue}{3.7\uparrow}}$ \\

    & PDBNet~\cite{dai2022pdbnet} & 2022 & $512 \times 1024$ & No & 1.82 & 15.0 & 2080Ti & 50 & $69.5^{ \color{blue}{3.5\uparrow}}$ \\

    & FPENet~\cite{liu2019feature} & 2019 & $512 \times 1024$ & No & 0.40 & 8.0 & 1080Ti & 55 & $70.1^{ \color{blue}{1.9\uparrow}}$ \\

    & AGFNet~\cite{zhao2022agfnet} & 2022 & $512 \times 1024$ & No & 1.12 & 14.4 & 2080Ti & 60 & $70.1^{ \color{blue}{1.9\uparrow}}$ \\

    & LRDNet~\cite{zhuang2021lrdnet} & 2021 & $512 \times 1024$ & No & 0.66 & 16.0 & 1080Ti & 77 & $70.1^{ \color{blue}{1.9\uparrow}}$ \\

    & CFPNet~\cite{luo2020latticenet} & 2021 & $1024 \times 2048$ & No & 0.55 & - & 1080Ti & 30 & $70.1^{ \color{blue}{1.9\uparrow}}$ \\

    & SFRSeg~\cite{singha2023real} & 2023 & $512 \times 1024$ & No & 1.60 & 19.6 & 2080Ti & 98 & $70.6^{ \color{blue}{1.4\uparrow}}$ \\

    & LEDNet~\cite{wang2019lednet} & 2019 & $512 \times 1024$ & No & 0.94 & 12.6 & 1080Ti & 78 & $70.6^{ \color{blue}{1.4\uparrow}}$ \\

    & SGCPNet~\cite{hao2022real} & 2022 & $1024 \times 2048$ & No & 0.61 & 4.5 & 1080Ti & 80 & $70.9^{ \color{blue}{1.1\uparrow}}$ \\

    & FBSNet~\cite{gao2022fbsnet} & 2023 & $512 \times 1024$ & No & 0.62 & 9.7 & 2080Ti & 90 & $70.9^{ \color{blue}{1.1\uparrow}}$ \\

    & EdgeNet~\cite{han2020using} & 2021 & $512 \times 1024$ & No & - & - & Titan X & 31 & $71.0^{ \color{blue}{1.0\uparrow}}$ \\

    & MSCFNet~\cite{gao2021mscfnet} & 2022 & $512 \times 1024$ & No & 1.15 & 17.1 & Titan XP & 50 & $71.9^{ \color{blue}{0.1\uparrow}}$ \\

     %& LMFFNet~\cite{shi2022lmffnet} & 2022 & $512 \times 1024$ & No & 1.4 & 16.7 & 2080Ti & \colorbox[RGB]{194,194,194}{118} & $75.1^{ \color{blue}{3.1\uparrow}}$ \\

     %& LCNet~\cite{shi2024lightweight} & 2024 & $512 \times 1024$ & No & 0.74 & 19.6 & 3090 & 117 & \colorbox[RGB]{183,183,183}{$75.6^{ \color{blue}{3.6\uparrow}}$} \\

     & LETNet~\cite{xu2023lightweight} & 2023 & $512 \times 1024$ & No & 0.95 & 13.6 & 2080Ti & \colorbox[RGB]{194,194,194}{150} &  \colorbox[RGB]{183,183,183}{$72.8^{ \color{red}{0.8\downarrow}}$} \\
    
\midrule
    & LMIINet (ours) & - & $512 \times 1024$ & No & 0.72 & 11.7 & 2080Ti & 100& 72.0 \\
\bottomrule
	\end{tabular}
\end{table*}

\begin{table*}[t]
	\setlength{\abovecaptionskip}{0pt}
	\setlength{\belowcaptionskip}{10pt}
	\caption{Comparisons with the state-of-art semantic segmentation methods on the CamVid test dataset.}
	\label{tab:cam}
        \centering
        \renewcommand\arraystretch{1.1}
	\begin{tabular}
 %{clcccccc}
 {p{2.4cm}p{1.6cm}<{\centering}p{1.8cm}<{\centering}p{1.8cm}<{\centering}p{1.6cm}<{\centering}p{2.0cm}<{\centering}p{1.6cm}<{\centering}p{1.6cm}<{\centering}}
\toprule
	Methods & Year & Resolution & Backbone & Param. (M)$\downarrow$
 & GPU & Speed (FPS)$\uparrow$ & mIoU ($\%$)$\uparrow$ \\
\midrule
	%ENet~\cite{paszke2016enet} & 2016 & $360 \times 640$ & No & \colorbox[RGB]{236,236,236}{0.36} &  Titan X & 135  & $51.30^{ \color{red}{18.64\downarrow}}$ \\
 
	%SegNet~\cite{badrinarayanan2017segnet} & 2017 & $360 \times 640$ & VGG-16 & 29.50 & Titan X & 15 & $55.60^{ \color{red}{14.34\downarrow}}$ \\
 
        NDNet~\cite{yang2020ndnet} & 2021 & $360 \times 480$ & - & 0.50 &  2080Ti & 50 & $57.20^{ \color{blue}{12.74\uparrow}}$ \\
        
        DFANet~\cite{li2019dfanet} & 2019 & $720 \times 960$ & Xception & 7.80 & Titan X & 120 & $64.70^{ \color{blue}{5.24\uparrow}}$ \\
        
        BiseNet-v1~\cite{yu2018bisenet} & 2018 & $720 \times 960$ & Xception & 5.80 &  Titan XP & 116 & $65.60^{ \color{blue}{4.34\uparrow}}$ \\
        
        DABNet~\cite{li2019dabnet} & 2019 & $360 \times 480$ & No & 0.76 & 1080Ti & 72 & $66.40^{ \color{blue}{3.54\uparrow}}$ \\
        
        FDDWNet~\cite{liu2020fddwnet} & 2020 & $360 \times 480$ & No & 0.80 & 2080Ti & 79 & $66.90^{ \color{blue}{3.04\uparrow}}$ \\
        
        ICNet~\cite{zhao2018icnet} & 2018 & $720 \times 960$ & PSPNet-50 & 26.50 & Titan X & 28 & $67.10^{ \color{blue}{2.84\uparrow}}$ \\

        PDBNet~\cite{dai2022pdbnet} & 2022 & $360 \times 480$ & No & 1.82 & 2080Ti & 90 & $67.70^{ \color{blue}{2.24\uparrow}}$ \\
        
        LBN-AA~\cite{dong2020real} & 2021 & $720 \times 960$ & No & 6.20 & 1080Ti & 39 & $68.00^{ \color{blue}{1.94\uparrow}}$ \\

        EFRNet-16~\cite{li2022efrnet}  & 2022 & $720 \times 960$   & EAA        & 1.44    & 2080Ti  & 154    & 
        $68.20^{ \color{blue}{1.74\uparrow}}$\\ 
        
        BiseNet-v2~\cite{yu2021bisenet} & 2020 & $720 \times 960$ & ResNet & 49.00 & 1080Ti & 60 & 
        $68.70^{ \color{blue}{1.24\uparrow}}$\\
        
        FBSNet~\cite{gao2022fbsnet} & 2023 & $360 \times 480$ & No & 0.62 & 2080Ti & 120 & $68.90^{ \color{blue}{1.04\uparrow}}$ \\
        
        SGCPNet~\cite{hao2022real} & 2022 & $720 \times 960$ & No & 0.61 & 1080Ti & \colorbox[RGB]{215,215,215}{278} & $69.00^{ \color{blue}{0.94\uparrow}}$ \\

        MLFNet~\cite{fan2022mlfnet}  & 2023 & $720 \times 960$   & ResNet-34        & 13.03    & Titan XP & 57 & 
        $69.00^{ \color{blue}{0.94\uparrow}}$\\ 

        LMFFNet~\cite{shi2022lmffnet} & 2023 & $360 \times 480$ & No & 1.40 & 3090 & 120 & $69.10^{ \color{blue}{0.84\uparrow}}$ \\
         
        MSCFNet~\cite{gao2021mscfnet} & 2022 & $360 \times 480$ & No & 1.15 &  Titan XP & 110 & $69.30^{ \color{blue}{0.64\uparrow}}$ \\

        AGFNet~\cite{zhao2022agfnet} & 2022 & $360 \times 480$ & No & 1.12 & 2080Ti & 100 & $69.30^{ \color{blue}{0.64\uparrow}}$ \\

        %LCNet~\cite{shi2024lightweight} & 2024 & $360 \times 480$ & No & 0.73 & 3090 & 140 & \colorbox[RGB]{194,194,194}{$70.30^{ \color{blue}{0.36\uparrow}}$} \\

        LETNet~\cite{xu2023lightweight} & 2023 & $360 \times 480$ & No & 0.95 & 2080Ti & 200 & \colorbox[RGB]{194,194,194}{$70.50^{ \color{red}{0.56\downarrow}}$} \\
        
\midrule

\multirow{2}{*}{LMIINet (ours)}
     & \multirow{2}{*}{-} & $360 \times 480$ & No & 0.72 & 2080Ti & 160 & 69.94 \\
    
    & & $720 \times 960$ & No & 0.72 & 2080Ti & 120 & 70.12 \\
\bottomrule
	\end{tabular}
\end{table*}

\subsection{Ablation Study}
\label{sec43}

To demonstrate the effectiveness of our network, we conducted a series of experiments using the CamVid dataset for both quantitative and qualitative assessments. We trained all model variations on the CamVid training set and assessed their performance on the CamVid validation set. Ablation studies were also carried out following the same methodology.

{\bfseries Improved Flatten Transformer:}
In this experimental section, we conducted comparative experiments involving various versions of the enhanced Flatten Transformer. Table~\ref{tab:Transformer} illustrates that without utilizing the Flatten Transformer, the segmentation accuracy remains at 68.79\% mIoU. However, with the inclusion of the Flatten Transformer, the model's performance improves to 69.01\% mIoU. Furthermore, by incorporating the CAB module and an attention module to create a dual-branch structure, the model's accuracy increases to 69.18\% mIoU. The highest segmentation accuracy of 69.94\% mIoU is achieved when implementing the feature interaction CC scheme in the dual-branch structure.

{\bfseries CC Scheme:}
In this experimental section, we conducted comparative experiments by adjusting the utilization and placement of the CC scheme. For the parameter in the CC module, the default value is set to 0.5. It is configured as a learnable parameter, allowing it to adapt and change dynamically with the training iterations. Table~\ref{tab:cc} shows that the model achieves a performance of only 68.60\% mIoU without CC usage. However, applying the CC scheme exclusively in the basic LFIB we designed enhances the model's performance by 0.58\%. Similarly, implementing the CC scheme exclusively in the dual-branch structure within the improved Flatten Transformer (consisting of the attention module and CAB module) results in a performance gain of 0.53\%. When the CC scheme is utilized in both LFIB and the improved Flatten Transformer, the model's performance gain increases to 1.34\%, demonstrating significant improvement.

{\bfseries SegHead:}
In our study, we conducted experiments using various combinations of outputs from the decoding stage as inputs to the semantic segmentation head. The results, presented in Table~\ref{tab:seghead}, reveal that the optimal performance was attained when the input combination was $\left\{1/2,1/4,1/8\right\}$.

{\bfseries LFIB:}
In this section, we conducted comparative experiments by adjusting the number of LFIBs at various stages. Regarding the number of layers in the LFIB module, our approach is as follows: First, during the encoding and decoding stages, we ensure that the number of LFIB blocks is consistent for features at the same resolution. Additionally, in the shallow stages, more spatial information can be captured, while in the deeper stages, more semantic information can be obtained. Therefore, we determine the number of LFIB blocks based on the above considerations, ensuring a balanced parameter count. As a result, the number of blocks in the deeper stages is typically greater than in the shallow stages. As depicted in Table~\ref{tab:nofLFIB}, our findings show that the segmentation accuracy of LMIINet peaks at 69.94\% mIoU when setting the number of LFIBs to 2 at each stage, while keeping the parameter count at a modest 720K. Furthermore, we observed that a more evenly distributed arrangement of blocks at each stage gradually enhances the model's performance.

{\bfseries Auxiliary Loss:}
Comparative experiments were conducted using features from various stages as inputs for the auxiliary loss function. Table~\ref{tab:Auxloss} shows that optimal performance is achieved when utilizing features downsampled to 1/8 as input. This results in optimal performance with minimal changes in parameters and computational load compared to other approaches.

\begin{figure*}[t]
        \centering
	\includegraphics[width=0.99\textwidth]{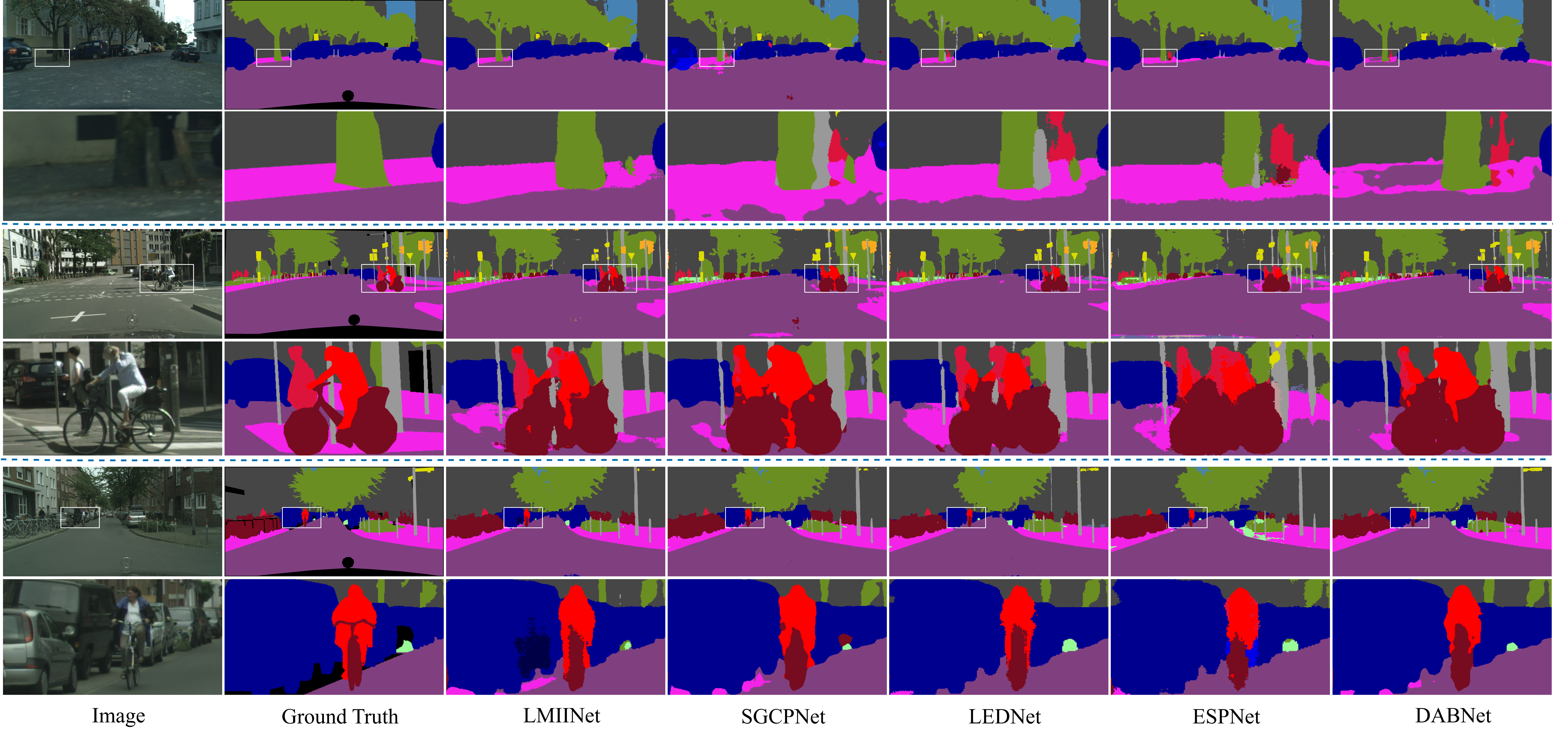}
	\caption{Visual comparisons on the Cityscapes dataset. From top to bottom are original input images, ground truths, and segmentation results from our \textbf{LMIINet}, SGCPNet~\cite{hao2022real}, LEDNet~\cite{wang2019lednet}, ESPNet~\cite{mehta2018espnet}, and DABNet~\cite{li2019dabnet}.}
	\vspace{-1.5em}
    \label{fig:city}
\end{figure*}

\begin{figure*}[t]
        \centering
	\includegraphics[width=0.99\textwidth]{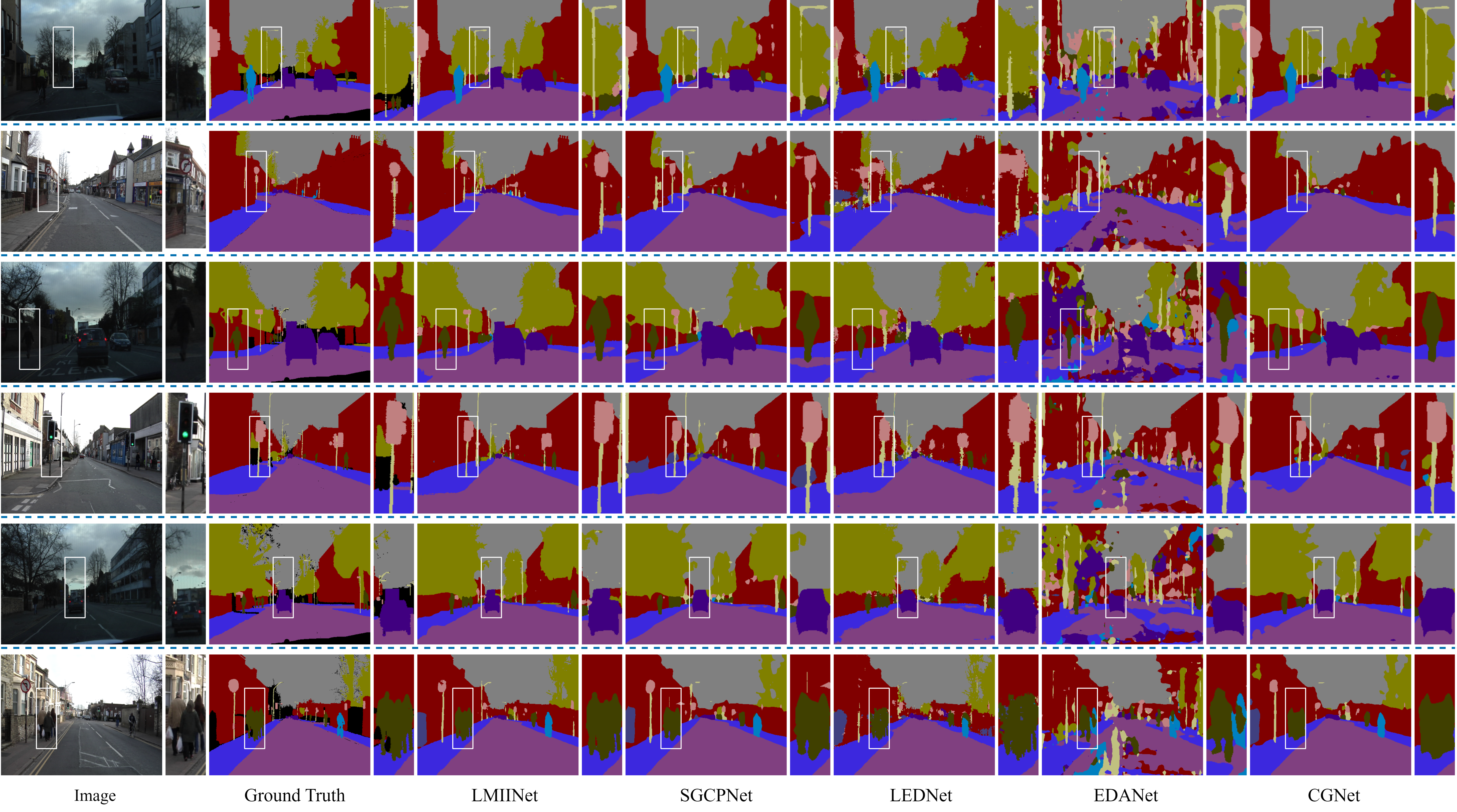}
	\caption{Visual comparisons on the CamVid dataset. From left to right are original input images, ground truths, and segmentation results from our \textbf{LMIINet}, SGCPNet~\cite{wang2019lednet}, LEDNet~\cite{wang2019lednet}, EDANet~\cite{lo2019efficient}, CGNet~\cite{wu2020cgnet}.}
	\vspace{-1.5em}
    \label{fig:cam}
\end{figure*}

\begin{table}[t]
	\setlength{\abovecaptionskip}{0pt}
	\setlength{\belowcaptionskip}{10pt}
	\caption{Comparisons of running speed of respective approaches on the Cityscapes test set by using RTX2080Ti with a resolution of $512 \times 1024$.}
	\vspace{-1em}
    \label{tab:speed}
        \begin{center}
 \renewcommand\arraystretch{1.1}
\begin{tabular}{p{1.4cm}p{1cm}<{\centering}p{0.95cm}<{\centering}p{0.95cm}<{\centering}p{0.95cm}<{\centering}p{0.95cm}<{\centering}}
\toprule
    Methods & Param. (M)$\downarrow$ & FLOPs (G)$\downarrow$ & Time (ms)$\downarrow$ & Speed (FPS)$\uparrow$ & mIoU ($\%$)$\uparrow$ \\   
\midrule
   ENet~\cite{paszke2016enet} & \colorbox[RGB]{215,215,215}{0.36} & \colorbox[RGB]{215,215,215}{2.4} & 13 & 
   80 & 57.8 \\
   
   NDNet~\cite{yang2020ndnet} & 0.50 & 10.0 & 17 & 60 & 64.2 \\

   CGNet~\cite{wu2020cgnet} & 0.50 & 9.0 & 78 & 12 & 66.0
   \\

  SCMNet~\cite{singha2021scmnet} & 1.20 & 18.2 & \colorbox[RGB]{215,215,215}{11} & \colorbox[RGB]{215,215,215}{92} & 69.5 \\

    PDBNet~\cite{dai2022pdbnet} & 1.82 & 15.0 & 20 & 50 & 69.5\\
\midrule
    AGFNet~\cite{zhao2022agfnet} & 1.12 & 14.4 & 17 & 60 & 70.1\\

    LRDNet~\cite{zhuang2021lrdnet} & 0.66 & 16.0 & 12 & 85 & 70.1\\

    SGCPNet~\cite{hao2022real} & \colorbox[RGB]{215,215,215}{0.61} & \colorbox[RGB]{215,215,215}{3.5} & 
    \colorbox[RGB]{215,215,215}{10} &  \colorbox[RGB]{215,215,215}{100} & 70.2\\

    SFRSeg~\cite{singha2023real} & 1.60 & 19.6 & 10 & 98 & 70.6\\

    LEDNet~\cite{wang2019lednet} & 0.94 &
    12.6 & 11 & 90 & 70.6\\
 
    FBSNet~\cite{gao2022fbsnet} & 0.62 & 9.7 & 11 & 90 & 70.9\\

    MSCFNet~\cite{gao2021mscfnet} & 1.15 & 17.1 & 17 & 60 & 71.9\\
\midrule
    \textbf{LMIINet} & 0.72 & 11.7 & \colorbox[RGB]{215,215,215}{10} & \colorbox[RGB]{215,215,215}{100} & \colorbox[RGB]{215,215,215}{72.0}\\
\bottomrule
\end{tabular}
\end{center}
\end{table}

\subsection{Comparisons with Advanced Models}
\label{sec44}
In this section, we compared our method with prominent semantic segmentation approaches on the Cityscapes and CamVid datasets. Our results show that our approach strikes a better balance between segmentation accuracy and efficiency.

\textbf{Evaluation on Cityscapes:}
As displayed in Table~\ref{tab:city}, we categorized existing methods into three groups based on parameters and computational complexity. Models with a parameter count below 2M are considered small size, while those with computational complexity exceeding 300G are deemed large size. Large-size methods prioritize segmentation performance at the expense of increased computational complexity, making them unsuitable for real-time processing on edge devices. In the medium-sized category, BiseNet-v2~\cite{yu2021bisenet} slightly outperforms our approach but has a parameter count five times larger than our LMIINet. LETNet~\cite{xu2023lightweight} outperforms our method in terms of performance, but our model has fewer parameters. Among small models, only MSCFNet~\cite{gao2021mscfnet} closely approaches our method, yet with a higher parameter count and complexity. LMIINet demonstrates favorable inference speed in small models, indicating a balanced performance and model size. %Additionally, the results in Table~\ref{tab:perclass} show LMIINet's superior performance in most classes. 
Visual comparisons with advanced semantic segmentation methods are presented in Fig.~\ref{fig:city}.

\textbf{Evaluation on CamVid:}
To verify our model's effectiveness and generalization capability, we compared it with other state-of-the-art methods on the CamVid dataset, detailed in Table~\ref{tab:cam}. AGFNet~\cite{zhao2022agfnet} closely matches our method but with a higher parameter count and complexity at the same input resolution. LMFFNet~\cite{shi2022lmffnet} performs worse than our method, despite having more parameters. LETNet~\cite{xu2023lightweight} outperforms our method, while it has more parameters. LMIINet maintains favorable inference speed, showcasing a balanced performance and model size. In contrast, SGCPNet~\cite{hao2022real} offers notable speed but lacks accuracy. Our LMIINet strikes a better balance between these aspects, delivering top performance with just 0.72M parameters. This underlines the effectiveness and superiority of the proposed LMIINet. Fig.~\ref{fig:cam} also shows the visual comparisons of advanced semantic segmentation methods.

\textbf{Speed Comparison:} 
To ensure a fair comparison, all methods were executed on the same platform to account for the impact of computational load on inference speed, which may vary based on the device used. Our evaluation was conducted on a single NVIDIA RTX 2080Ti GPU to measure model execution times, maintaining a spatial resolution of  $512 \times 1024$ during the experiments. The comparison of speed and runtime between our proposed LMIINet and other lightweight methods is presented in Table~\ref{tab:speed}. The results showcase LMIINet's impressive speed, achieving a frame rate of 100 FPS for processing image streams of size $512 \times 1024$, establishing it as the fastest method. Despite ENet~\cite{paszke2016enet} and SGCPNet~\cite{hao2022real} having smaller parameter counts and computational costs, LMIINet's competitive accuracy of 72.0\% and inference speed of 100 FPS is pivotal for practical applications like autonomous driving. With an effective balance of speed and accuracy, LMIINet stands out as a compelling choice for real-world deployment.

\section{CONCLUSION}
\label{sec:conclusion}

This paper presents LMIINet, a real-time semantic segmentation model based on a lightweight multiple-information interaction network. It combines CNNs with Transformers to leverage local and global feature extraction capabilities. The improved Flatten Transformer enhances performance by enabling interaction between the self-attention module and CNN features. The basic LFIB module facilitates effective feature interaction while maintaining simplicity and lightness. By incorporating the CC learning scheme and a dual-branch structure, the model achieves high-level information interaction, significantly improving performance. Experimental results confirm the model's balance between parameters and performance, showcasing the effectiveness of the feature interaction strategy in achieving satisfactory performance under lightweight conditions.

\bibliographystyle{IEEEtran}
\bibliography{reference}

\end{document}